\documentclass[conference]{IEEEtran}
\IEEEoverridecommandlockouts
\usepackage{cite}
\usepackage{amsmath,amssymb,amsfonts}
\usepackage{xspace, amsthm, bbm, bm,theoremref}
\usepackage{paralist}
\usepackage{todonotes}\usepackage{algorithmic}
\usepackage{graphicx}
\usepackage{textcomp}
\usepackage{xcolor}
\usepackage{thmtools,thm-restate}
\usepackage{algorithm2e}
\usepackage{algorithmic}
\usepackage{multirow}
\usepackage[caption=false]{subfig} 
\usepackage{authblk}
\newtheorem{definition}{Definition}

\renewcommand{\P}[2]{\mathbf{P}_{#1}\left[#2\right]}

\def\BibTeX{{\rm B\kern-.05em{\sc i\kern-.025em b}\kern-.08em
    T\kern-.1667em\lower.7ex\hbox{E}\kern-.125emX}}

\newtheorem{observation}{Observation}
\DeclareMathOperator*{\argmin}{arg\,min}
\begin{document}

\title{Collaborative and Privacy-Preserving Machine Teaching via Consensus Optimization} 
\author[1]{Yufei Han}
\author[2]{Yuzhe Ma}
\author[1]{Christopher Gates}
\author[1]{Kevin Roundy}
\author[1]{Yun Shen}
\affil[1]{Symantec Research Labs}
\affil[2]{University of Wisconsin-Madison}

\maketitle

\begin{abstract}

In this work, we define a collaborative and privacy-preserving machine teaching paradigm with multiple distributed teachers. We focus on consensus super teaching. It aims at organizing distributed teachers to jointly select a compact while informative training subset from data hosted by the teachers to make a learner learn better. The challenges arise from three perspectives. First, the state-of-the-art pool-based super teaching method applies mixed-integer non-linear programming (MINLP) which does not scale well to very large data sets. Second, it is desirable to restrict data access of the teachers to only their own data during the collaboration stage to mitigate privacy leaks. Finally, the teaching collaboration should be communication-efficient since large communication overheads can cause synchronization delays between teachers. 


To address these challenges, we formulate the collaborative teaching as a consensus and privacy-preserving optimization process to minimize teaching risk. We theoretically demonstrate the necessity of collaboration between teachers for improving the learner's learning. Furthermore, we show that the proposed method enjoys a similar property as the Oracle property of adaptive Lasso. Empirical study illustrates that our teaching method can deliver significantly more accurate teaching results with high speed, while the non-collaborative MINLP-based super teaching becomes prohibitively expensive to compute. 

\end{abstract}



\section{Introduction}
Machine teaching \cite{goldman1995complexity,shinohara1991teachability,zilles2011models,zhu2013machine,chen2018understanding} studies the inverse problem of machine learning, where a teacher already has a specific target model ($\theta^{*}$) it wants to teach some other student (learner), and the teacher designs the optimal training set such that the student can efficiently learn the target model.
The constructed training set does not need to be independent and identically 
distributed. The teacher is allowed to design any instance in the input space, which enables flexibility when generating an efficient training set. The efficacy of teaching is 
measured by computational cost of the model training, accuracy of the derived learner, robustness of the training process and so on. 
In general, machine teaching places a strong emphasis
on the teacher and its power to control data. Machine teaching is connected to machine learning fundamentals as it defines abstractions and interfaces between the learning algorithm and the teaching paradigm. 
Research on machine teaching has not only great theoretical value, but also applications in personalized education and human-in-the-loop learning. 

Super teaching is an interesting phenomenon unveiled recently in machine teaching. As stated in~\cite{ma2018teacher}, a learner is super-teachable if a teacher can
trim down an \textit{\textit{i.i.d.}} training set while enhancing the learning performance. Distinct from training set reduction where the target model is hidden from the learner, super teaching assumes the teacher knows the target model and rely on such knowledge to select a training subset so that a student learner can learn better on that subset.

Prior work in super teaching assumes that only one central teacher is present and it has full knowledge of all data instances used for teaching. As the privacy concern of data security becomes increasingly important, research in super teaching faces the following important challenges. First, instead of transferring all data used to teach to the central teacher where teaching is conducted, it is preferable to keep data on the local devices and conduct teaching with multiple distributed teachers. Each teacher only accesses the data samples hosted by itself. It avoids heavy overheads of the data transferring and prevents leaks of private information contained in the local data set. However, it is not clear whether organizing a consensus collaboration between teachers can provide merits of teaching compared to independently conducting teaching by each teacher in a stand-alone mode. Little efforts have been devoted to discuss how to organize an efficient collaborative super teaching paradigm to achieve good teaching performances and a privacy-preserving process of teaching at the same time. 
Furthermore, the state-of-the-art super teaching method proposed in~\cite{ma2018teacher} is formulated as a mixed-integer non-linear programming (MINLP) problem. 
The general computational complexity of MINLP problem is undecidable in theory~\cite{leoliberti2019}. In the worst case, the popular heuristic solver, such as Branch-and-Bound (BnB) method, has an exponential time complexity thus becomes prohibitively expensive given large-scale training data. Solving a MINLP problem with distributed players is even more difficult, as it usually needs a central processor to allocate the resource across multiple players to solve the overall problem \cite{EzgiGT2017}. Therefore, the central processor can access the local private data, which potentially violates the privacy regulation. Besides, frequent synchronization between the central process and the end-devices can cause severe latency given a low-communication environment. Finally, limited computing capability of end-devices in a distributed network can not afford to the intensive computation of solving the MINLP problem. 

We propose a novel computationally efficient distributed super teaching paradigm, which coherently facilitates collaboration between multiple teachers in a privacy-preserving way. Our study confirms a well-known intuition: \textbf{\textit{A carefully organized consensus collaboration between different teachers can enhance the teaching performances}}. We also show that independently conducting teaching in a non-colluded way can even make the teaching performance deteriorate. Furthermore, the privacy-preserving design of the proposed collaborative teaching paradigm encourages information sharing between teachers in the collaboration stage. 

\section{Related work}\label{sec:related_work}

\subsection{Machine Teaching}
Machine teaching was originally proposed in~\cite{goldman1995complexity, shinohara1991teachability}. It has attracted plenty of research interest, most of which focus on studying a  quantity called the teaching dimension, i.e., the size of the minimal training set that is guaranteed to teach a target model to the student. For example,~\cite{goldman1995complexity} provides a discussion on the teaching dimension of version space learners,~\cite{liu2016teaching} analyzes the teaching dimension of linear classifiers, and~\cite{zhu2013machine} studies the optimal teaching problem of Bayesian learners. In standard machine teaching, the student is assumed to passively receive the optimal training set from teacher. Later works consider other variants of teaching setting, e.g., in~\cite{zilles2011models,balbach2008measuring}, the student and the teacher are allowed to cooperate in order to achieve better teaching performance. More studies of machine teaching can be found in~\cite{doliwa2014recursive,chen2018understanding,haug2018teaching,liu2017towards,liu2017iterative}. 

Machine teaching as a theoretical regime also has many applications in cognitive science and computer security. One application is the personalized education, where a clairvoyant teacher can help design minimal teaching curriculum for a  human student such that an educational goal is achieved~\cite{patil2014optimal}. As another popular application, machine teaching can be used to perform data poisoning attacks of real-world machine learning systems~\cite{mei2015using,alfeld2016data,ma2018data}. In such cases, the teacher is viewed as a nefarious attacker who has a specific attack goal in his mind, while the student is some machine learner, then the teaching procedure corresponds to minimally tampering the clean dataset such that the machine learner learns some sub-optimal target model on the corrupted training set.


Instead of artificially designing the training set, super teaching ~\cite{ma2018teacher} selects a subset from an \textit{i.i.d.} training set to conduct teaching. 
Mathematically, super teaching is defined as below.
\begin{definition}[Super Teaching]
\label{def:superteaching}
Let $S$ be an $n$-item \textit{iid} training set, and $T$ be a teacher who selects a subset $T(S)\subset S$ as the training subset for learner $A$. Let $\hat\theta_S$ and $\hat\theta_{T(S)}$ be the model learned from $S$ and $T(S)$ respectively. Then $T$ is a super teacher for learner $A$ if~$\forall\delta>0, \exists N$ such that $\forall n\ge N$
\begin{equation}
\small
\P{S}{R(\hat\theta_{T(S)}) \le c_n R(\hat\theta_S)} > 1-\delta,
\end{equation}
where $R$ is some teaching risk function, the probability is with respect to the randomness of $S$, and $c_n \le 1$ is a sequence called the super teaching ratio.
\end{definition}

The idea of selecting an informative training subset is also explored in \cite{Yangfan2018}. In the proposed \textit{learning-to-teach} framework, the teacher conducting subset selection is modeled with Deep Neural Nets (DNN). The goal of teaching is to select training samples to make faster convergence of the DNN based learner. The teacher network is tuned via reinforcement learning with reward signals encouraging fast descent of the classification loss of the learner. In contrast to learning-to-teach, super teaching in~\cite{ma2018teacher} focuses on a more general teaching goal, which drives the student to learn the expected model. 
Although only simple learners such as Logistic Regression are considered in~\cite{ma2018teacher}, the theoretical study over the teachability of super teaching can be further applied to many advanced learners. 

Inspired by the teachability theory proposed in \cite{ma2018teacher}, our work extends the horizon of super teaching by studying applicability of a collaborative and privacy-preserving teaching scenario. Different from learning-to-teach, multiple teachers are present as collaborative players in the teaching activity. Furthermore, training data hosted by any one teacher can not be accessed by the others in our problem setting, whereas learning-to-teach assumes the teacher network can access all the training data. 

\subsection{Federated Learning}
Another relevant branch of research is federated learning \cite{Konecny2016}. Federated learning is a communication-efficient and privacy-preserving distributed model training method over distributed agents.
Each agent hosts their own data instances and is capable of computing local model update. In each round of model training, the training process is first conducted on each node in parallel without inter-node communication.  
Only the local model updates are aggregated on a centralized parameter server to derive the global model update. The aggregation is agnostic to data distribution of different agents. Neither the centralized server, nor the local agents have visibility of the data owned by any specific agent. In \cite{MJaggi2014},  a communication-efficient distributed optimization method named \textit{CoCoA} is proposed for training models in a privacy-preserving  way. \textit{CoCoA} applies block-coordinate descent over the dual form of the joint convex learning objective and guarantees sub-linear convergence of the federated optimization. Furthermore, the optimization process does not require to access data instances hosted by each node. Only local dual variable updates need to transfer from local nodes to the central server. This property makes \textit{CoCoA} inherently appropriate for federated training. 

A federated data poisoning attack is recently proposed in \cite{Bhagoji2018}. This work assumes that only one malicious agent conducts non-colluding adversarial data poisoning over the local data instances that it hosts. 
Our method is distinct from this work since we study consensus collaboration of multiple teachers. 
In addition, we investigate a more generous goal of teaching than data a pre-specified target model with that training set., which guides the learner to learn a pre-specified yet potentially malicious target model. 

\section{Collaborative Super Teaching}\label{sec:fed_teacher}
We assume $K$ teachers and one central parameter server as the learner. Each teacher hosts a local private dataset $D_i$ ($i\in[K]$) of size $N_i$. As the output of super teaching, each teacher selects a subset $S_i\subset D_i$. The learner runs the learning algorithm $L$ on the aggregated subsets $S=\cup_{i\in[K]}S_i$ to obtain the model. Each teacher only accesses its own data $D_i$ during the teaching process due to the privacy-preserving regulations.
Once the teaching stage terminates, the learner can further conduct federated model training to keep the local data subsets $\{S_i\}$ on the local machines, which protects teachers' data privacy after teaching.  
Discussing how to conduct model training is beyond our scope. Without loss of generality, we assume that the learner choses federated training. 

We set the teaching goal as the value of the model parameter $\theta^{*}$ that the teachers expect the learner to obtain, as the setting in~\cite{ma2018teacher}.  In the collaborative environment, the union of the selected subsets $\{S_i\}$ should be jointly helpful in inducing $\theta^{*}$. 
Therefore we propose to define the collaborative super teaching as in~\eqref{eq:primal_super_teacher}
\begin{equation}\label{eq:primal_super_teacher}
\begin{split}
&\hat\theta_{S}, \,\, b^i:i\in[K] = \underset{\hat\theta_{S}, b^{i}:i\in[K]} \argmin {R^{*}(\hat\theta_S)}\\
s.t. & \,\,\,\, \hat\theta_{S} = \underset{\theta} \argmin \sum_{i=1}^{K} \sum_{j=1}^{N_i} b^i_{j} \ell(\theta, x^{i}_{j},y^{i}_{j}) + \frac{\lambda}{2}\Omega(\theta), \\
& \,\,\,\, b^i\in \{0,1\}^{N_i}, \forall i\in [K],\\
\end{split} 
\end{equation}
where $R^*(\hat\theta_S) = \|{\hat\theta_S} - {\theta^{*}}\|$ measures the teaching risk as Euclidean distance between $\hat\theta_S$ and $\theta^{*}$, $(x^{i}_j, y^{i}_j)$ is the $j$th data instances of $D_{i}$ hosted by the teacher $i$, and $b^i$ is an $N_i$-dimensional binary-valued vector with $b^i_j = 1$ denoting the instance $(x^{i}_j, y^{i}_j)$ is selected and $b^i_j = 0$ otherwise.  $\ell$ is the learning loss function, $\Omega(\theta)$ is the regularization over the model complexity of the learner and $\lambda$ is the regularization weight. 
Intuitively, there is a primitive solution to the proposed distributed teaching problem: \textbf{oblivious teaching}, where each teacher independently selects its own teaching set without collaborating with the other teachers. The independently selected subsets are aggregated to form the training set of the learner. The questions of interest are thus i) whether the oblivious teaching can reduce the teaching risk. and ii) whether it is possible to improve teaching performance by organising appropriate collaboration between the teachers, compared to the oblivious teaching. 

For simplicity of analysis, we assume that $\Omega(\theta)$ is the $l_2$-norm penalty $\|\theta\|^2$ and the loss $\ell$ is a convex and $\tau$-smooth function of $\theta$, which holds in many cases such as the logistic loss or squared loss with bounded input space.
Thus, the learning algorithm $L$ of the learner takes the form of a convex optimization. Based on such assumption, we provide an initial answer to the question i) in Theorem~\ref{thm:kparty}:
\begin{restatable}{theorem}{thm-kparty}
\label{thm:kparty}
Assume the model space $\Theta$ is bounded, i.e., $\forall\theta\in \Theta$, $\|\theta\|\le B$. Also assume the convex learning loss $\ell$ is $\tau$-smooth, i.e., $\|\nabla\ell(\theta)-\nabla\ell(\theta^\prime)\|\le \tau \|\theta-\theta^\prime\|, \forall \theta,\theta^\prime$. The teaching risk is defined as $R^*(\hat \theta)=\frac{1}{2}\|\hat\theta-\theta^*\|^2$, where $\theta^*$ is the target model. Each teacher independently solves~\eqref{eq:primal_super_teacher} on its own dataset $D_i$ with regularization weight $\frac{\lambda}{K}$, and let $S=\cup_{i\in [K]}S_i$ be the aggregated dataset. Then 
\begin{equation}
R^*(\hat\theta_S)\le (\frac{\tau}{\lambda K}+\frac{1}{K^2})\sum_{i=1}^KR^*(\hat\theta_{S_i}).
\end{equation}
\end{restatable}
\begin{proof}
Let the $(\hat x_j^i,\hat y_j^i)$ be the $j$th point in the selected $S_i$.
Define $g_i(\theta)=\sum_{j=1}^{|S_i|} \ell(\theta, \hat x_j^i, \hat y_j^i)+\frac{\lambda}{2K}\| \theta\|^2$, where $|S_i|$ is the number of items in $S_i$. Then $\hat\theta_{S_i}=\argmin_{\theta\in\Theta} g_i(\theta)$. Define 
\begin{equation}
g(\theta)=\frac{1}{K}\sum_{i=1}^K g_i(\theta)=\frac{1}{K}(\sum_{i=1}^K\sum_{j=1}^{|S_i|}\ell(\theta, \hat x_j^i,\hat y_j^i) +\frac{\lambda}{2}\|\theta\|^2).
\end{equation}
Then $\hat\theta_S=\argmin_{\theta\in\Theta}g(\theta)$. Since $g(\theta)$ is $\lambda$-strongly convex, 
\begin{equation}
g(\theta^*)-g(\hat\theta_S)\ge\frac{ \lambda}{2}\|\theta^*-\hat\theta_S\|^2=\lambda R^*(\hat\theta_S).
\end{equation}
Thus $R^*(\hat\theta_S)\le \frac{1}{\lambda}(g(\theta^*)-g(\hat\theta_S))$. 
Next we upper bound $g(\theta^*)-g(\hat\theta_S)$. Note that $g(\theta^*)-g(\hat\theta_S)=$
\begin{equation}\label{eq:bound}
\frac{1}{K}\sum_{i=1}^K (g_i(\theta^*)-g_i(\hat\theta_S))\le\frac{1}{K}\sum_{i=1}^K (g_i(\theta^*)-g_i(\hat\theta_{S_i})),
\end{equation}
where the last inequality is due to $\hat\theta_{S_i}=\argmin_{\theta\in\Theta}g_i(\theta)$. 
Since the loss $\ell$ is $\tau$-smooth, one can easily show that $g_i(\theta)$ is $(\tau+\frac{\lambda}{K})$-smooth, thus we have the following upper bound
\begin{equation}\label{eq:bound2}
\small
\begin{aligned}
g_i(\theta^*)-g_i(\hat\theta_{S_i})&\le \frac{1}{2}(\tau+\frac{\lambda}{K})\|\theta^*-\hat\theta_{S_i}\|^2=(\tau+\frac{\lambda}{K})R^*(\hat\theta_{S_i}).
\end{aligned}
\end{equation}
By~\eqref{eq:bound} and~\eqref{eq:bound2}, we have
\begin{equation}
R^*(\hat\theta_S)\le \frac{1}{\lambda}(g(\theta^*)-g(\hat\theta_S))\le (\frac{\tau}{\lambda K}+\frac{1}{K^2})\sum_{i=1}^KR^*(\hat\theta_{S_i}).
\end{equation}
\end{proof}
Theorem~\ref{thm:kparty} implies that if $\forall i$, $R^*(\hat\theta_{S_i})\le c$, then $R^*(\hat\theta_S)\le(\frac{\tau}{\lambda}+\frac{1}{K})c$, which is a constant change. Thus by simply aggregating the selected subsets, one can achieve comparable teaching performance as each individual teacher does. However, the joint teaching performance is never guaranteed to be better than individual teaching. 
Therefore it is obvious that in order to achieve better joint teaching performance, the teachers should share information with each other and decide how to tweak their own teaching subset based on the selection made by the peers. This is also intuitive in real-world education, where human teachers collaboratively teach student better via communication with each other. Motivated by this observation, we define a collaborative teaching strategy that encourages information sharing between teachers to jointly minimize the teaching risk, while keeping the local data private.

\subsection{Regularized Dual Learning for Collaborative Teaching}
The dual objective of the learning paradigm for the learner gives: 
\begin{equation}\label{eq:dual_learner}
\alpha^{*} = \underset{\alpha}{\argmin} \sum_{i=1}^{K} \sum_{j=1}^{N_i} \ell^{*}(-\alpha^{i}_j) + \frac{\lambda}{2} \|Z\alpha\|^2 
\end{equation}
where $\ell^{*}$ is the Fenchel dual of the loss function $\ell$. Let $N = \sum_{i=1}^{K} N_i$ denote the number of training instances delivered by the teachers. $Z \in R^{d*N}$ denotes aggregated data matrix with each column corresponding to a data instance. 
The duality comes with the mapping from dual to primal variable: $\omega(\alpha) = Z\alpha$ as given by the KKT optimality condition.    
$\alpha$ is the $N$-dimensional dual variable, where each $\alpha^{i}_j$ denotes the dual variable corresponding to the $j$th data instance hosted by teacher $i$. If $\alpha^{i}_{j}$ diminishes, the corresponding data instance $Z^{i}_{j}$ consequently has no impact over the dual objective in~\eqref{eq:dual_learner}. Thus, only the data instances with non-zero $\alpha^{i}_j$ dominates the training process. Motivated by this observation, we propose to optimize the dual objective and enforce sparsity structure of $\alpha$ simultaneously to achieve selection of the informative training samples in~\eqref{lr_dual_augmented}. Bearing in mind the goal of the collaborative teaching, we also introduce an additional quadratic penalty shrinking the gap between $\theta^{*}$ and the learnt model $Z\alpha$. 
\begin{equation}\label{lr_dual_augmented}
\begin{split}
\alpha  &= \underset{\alpha^{i}_{j},i\in [K]} {\argmin} \frac{1}{n}\sum_{i=1}^{K}\sum_{j=1}^{N_i} \ell^{*}(-\alpha^{i}_j) + \frac{\lambda}{2}\|Z\alpha\|^2 \\
&+ \lambda_{\theta} \|\theta^{*} - Z\alpha\|^2  +   \lambda_{\alpha} \sum_{i=1}^{K}\sum_{j=1}^{N_i} w^{i}_j |\alpha^{i}_j|
\end{split}
\end{equation}
where $\lambda_{\alpha}$ and $\lambda_{\theta}$ are the weight coefficients of the adaptive $l_1$-norm based penalization enforcing sparsity of $\alpha$ and the quadratic penalty minimizing the teaching risk $R^{*}$.  $w^{i}_{j}$ is data-dependent per-variable weight assigned to each dual variable $\alpha^{i}_{j}$. Based on \cite{HZhou2006}, $w^{i}_{j}$ can be set up as $1/|{\hat\alpha}^i_j|$. $1/|{\hat\alpha}^i_j|$ denotes a warm-start estimate of $\alpha^{i}_j$, which can be derived by simply calculating the Ordinary-Least-Squares solution to $\|\theta^{*} - Z\alpha\|^2$. The teaching objective given in~\eqref{lr_dual_augmented} is apparently convex according to the property of Legendre-Fenchel transform. Thus solving~\eqref{lr_dual_augmented} with gradient descent guarantees fact convergence. As enforced by the $l_1$-norm regularization over $\alpha$, the non-zero entries of the optimal $\alpha$ of the objective function in~\eqref{lr_dual_augmented} correspond to the most useful data instances for the learner to reach the expected teaching goal and minimize the teaching risk $R^{*}$. In practice, the learned $\alpha$ has a small fraction of entries with dominant magnitudes, and rest are negligible. 
We thus rank the data instances $Z^i_{j}$ according to the magnitude of $|\alpha^i_{j}|$. The top-ranked $|S|$ data instances with the largest $|\alpha^i_j|$ are selected to form the final training subset for the learner. Since the selected data instances are distributed over different teachers. Solving~\eqref{lr_dual_augmented} helps to jointly identify which data instances on each of the $K$ teachers should be used to teach the learner. In the consensus optimization, each teacher learns to conduct the selection based on the decisions of the other teachers. Compared to heuristically tuning each teacher's decision, solving~\eqref{lr_dual_augmented} explicitly coordinates different teachers to deliver collaborative subset selection to minimize the teaching risk globally. Furthermore, we observe that the solution to~\eqref{lr_dual_augmented} enjoys a similar property as the Oracle property of adaptive Lasso~\cite{HZhou2006} with an appropriately chosen $\lambda_{\theta}$ and $\lambda_{\alpha}$, as given in Observation \ref{obs:lasso}.
\begin{observation}\label{obs:lasso}
Given a training set $\{(x_i,y_i)\}_{i\in [N]}$, where $x_i\in \mathcal{R}^d$. We assume that the goal of the super teaching satisfies $\theta^{*}=\sum_{i\in A}\alpha_i^*x_i$, where $A\subset [N]$,
and $\alpha^{*}$ denotes the dual variable. We further assume that $\lambda_{\theta} \gg \varphi$ where $\varphi$ is the empirical upper limit for the learner's classification loss on the training set. Given $\gamma > 0$, if $\frac{\lambda_{\alpha}}{\lambda_{\theta}}d^{-\frac{1}{2}} \to {0}$ and  $\frac{\lambda_{\alpha}}{\lambda_{\theta}}d^{-\frac{1}{2}+\frac{\gamma}{2}}\to {\infty}$ (see Theorem 2 in \cite{HZhou2006}), then the global optimal solution $\alpha$ to~\eqref{lr_dual_augmented} must satisfy the Oracle property: $\lim_{d\rightarrow \infty} P(\alpha = \alpha^*) = 1$. 
\end{observation}



\subsection{privacy-preserving  teaching via block-coordinate descent}
We propose to use block-coordinate descent to solve~\eqref{lr_dual_augmented}. In each round of the descent process, we minimize~\eqref{lr_dual_augmented} with respect to all $\alpha^{i}_{j}$ belonging to the same teacher $i$, while fixing all the other $\alpha$ as constants. The pseudo codes of the optimization procedure is given in algorithm~\ref{alg:fedML}.

\begin{algorithm}[t]
\caption{Block-Coordinate Descent for Collaborative Super Teaching}\label{alg:fedML}
\KwData{$\{z^{i}_j \, \, i = 1,2,3...,K, j = 1,2,3,...,N_i\}$ hosted by $K$ teachers} 
\KwIn{$T \geq 1$ as the maximum iteration steps, scaling parameter $1 \leq \beta_{i} \leq {K}$, by default $\beta_{i} = 1$} 
\KwOut{${\alpha^{i}_{j}},i=1,2,...,K,j=1,2,...,N_i$}
\textbf{Initialize}: $\alpha^{i}_{j} = 0$ for all machines and ${\widetilde\theta}^{(0)} = 0$\\
\For{$t=1$ \KwTo $T$}{ 
      \For{\textit{all teachers} $i=1,2,3,...,K$ in parallel}{
             $\Delta {\alpha}^{i} = \underset{\Delta{\alpha}^{i}}{\argmin} \frac{\lambda}{2}\|{\widetilde\theta}^{(t-1)} + \frac{1}{\lambda}\sum_{j=1}^{N_i}\Delta{\alpha}^{i}_j x_j^i y_j^i\|^2 + {\ell}^{*}(-\alpha^{(t-1),i} - \Delta{\alpha}^{i}) + \sum_{j=1}^{N_i} w^{i}_{j} |\alpha^{i}_j+ \Delta{\alpha}^{i}_j| + \lambda_{\theta}\|{\widetilde\theta}^{t-1} +\frac{1}{\lambda}\sum_{j=1}^{N_i}\Delta{\alpha}^{i}_{j} x^{i}_j y^{i}_j - \theta^{*}\|^2$\\
             $\alpha^{t,i} = \alpha^{t-1,i} + \frac{\beta_{i}}{K}\Delta{\alpha}^{i}$\\
      }
      
      \textit{Reduce on the central parameter server} ${\widetilde\theta}^{t} = {\widetilde\theta}^{t-1} + \frac{1}{\lambda}\sum_{i=1}^{K}\sum_{j=1}^{N_i} \alpha^{t,i}_{j} x^{i}_j$
      
      \textit{Broadcast $\widetilde\theta^{t}$ to all $K$ teachers}
    }
\end{algorithm}
We use $\alpha^{t,i}$ to denote the disjoint block $\{\alpha^{i}_j\},j=1,2,3..,N_i$ corresponding to the data instances hosted by teacher $i$, which are estimated at the $t$-th iteration. $Z^{i}$ denotes the columns in the data matrix $Z$ storing the data instances of the teacher $i$.  In each round of iteration, we update the dual variable $\alpha^{i}$ for each of the $K$ teachers in parallel. We assume an incremental update $\Delta\alpha^{i}$ based on the value of $\alpha^{t-1,i}$. This incremental variation indicates the descent direction minimizing the teaching loss with respect to the block $\alpha^{i}$. It is estimated by minimizing the local approximation to~\eqref{lr_dual_augmented}, where $\alpha^{i}$ is represented as the additional combination $\alpha^{t-1,i} + \Delta\alpha^i$. $\beta_{i}$ is the learning rate adjusting the descent step length for the block $\alpha^{i}$. Note updating each block $\alpha^{i}$ does not require knowledge of the values for the other blocks. All the local updates need is the local dual variable value $\alpha^{t-1,i}$ obtained from the last round and the global aggregated variable $\widetilde{\theta}$ broadcasted from the central server. As such, update of each block can be conducted in parallel without inter-teacher communication. Similarly, aggregating to derive the global variable $\widetilde{\theta}$ is also a parallel process. The teachers forward the local aggregation $\sum_{j=1}^{N_i} \alpha^{t,i}_{j} x^{i}_j$ to the central server, where simply summing up the local aggregation gives the global variable value.
It is worth noting that we use $\widetilde{\theta}$ to denote the global aggregation variable. It does not imply the primal-dual correspondence, as we are solving a different problem from~\eqref{lr_dual_augmented}. Throughout the block-coordinate descent process, it is easy to find that 
i)  private data hosted by any teacher is kept on its own machine in the collaboration stage. In other words, no training data is transferred directly between teachers. Furthermore, updating $\widetilde{\theta}$ only needs to transfer the local aggregation $\sum_{j=1}^{N_i} \alpha^{t,i}_{j} x^{i}_j$ to the central sever. It is difficult to infer any statistical profiles about the local data of the teachers based on solely on the local aggregation $\sum_{j=1}^{N_i} \alpha^{t,i}_{j} x^{i}_j$, which reduces the risk of unveiling local private data of one teacher to the others in the collaboration step. 
ii) sharing information between different teachers is conducted in the proposed method by updating the global aggregation variable $\widetilde{\theta}$ and then broadcasting the updated value to all $K$ teachers. Communication for teaching collaboration is thus efficient, with the cost of $O(Kd)$ in each round of iteration. Moreover, according to \cite{MJaggi2014}, updating $\alpha^{i}$ of local teachers can be triggered with asynchronous parallelism, which allows to organize efficient teaching collaboration with large number of teachers and tight communication budget. 


We demonstrate how to apply the proposed super teaching method to two prevalent learners, $l_2$-regularized Logistic Regression (LR) and Ridge Regression (RR). 
\subsubsection{Collaborative Teaching for $l_2$-regularized Logistic Regression}
$(x^{i}_j, y^{i}_j)$, $i=1,2,3,..,K, j = 1,2,3,..,N_i$ denote the features and labels of the data instances hosted by all $K$ teachers. To instantiate~\eqref{lr_dual_augmented} to $l_2$-regularized Logistic Regression, we concretize the definition of $\ell^{*}$ with slight modification on the weight parameters, which gives:  
\begin{equation}\label{lr_dual_regularized}
\small
\begin{split}
\alpha &= \underset{\alpha}{\argmin} \frac{\lambda}{2} \sum_{i=1}^{K} \sum_{j=1}^{N_i}\|\frac{1}{\lambda}\alpha^{i}_{j} x^{i}_j y^{i}_j\|^2 + \sum_{i=1}^{K} \sum_{j=1}^{N_i} \ell^{*}(-\alpha^{i}_j) \\
&+ \lambda_{\alpha} \sum_{i=1}^{K} \sum_{j=1}^{N_i} w^{i}_{j} |\alpha^{i}_j| + \lambda_{\theta} \|\theta^{*} - \frac{1}{\lambda{n}}\sum_{i=1}^{K}\sum_{j=1}^{N_i} \alpha^{i}_j y^{i}_{j} x^{i}_j\|^2\\
&\mbox{s.t.} 0 \leq \alpha^{i}_{j} \leq 1\\
\end{split}
\end{equation}
where $y^{i}_j$ is the binary class label of the data instance, valued as +/-1 and $\ell^{*}(-\alpha^{i}_j) = \alpha_i{\log{\alpha_i}} +(1-\alpha_i)\log(1-\alpha_i)$ 

The collaborative super teaching for LR is defined as a box-constrained convex quadratic programming problem following the principle of algorithm~\ref{alg:fedML}. The optimization process is given in algorithm~\ref{alg:fedML_lr}:
\begin{algorithm}[t]
\caption{Block-Coordinate Descent for Collaborative Super Teaching of $l_2$-Regularized Logistic Regression}\label{alg:fedML_lr}
\textbf{Initialize}: $\alpha^{i}_{j} = 0$ for all teachers and ${\widetilde\theta}^{(0)} = 0$\\
\For{$t=1$ \KwTo $T$}{ 
      \For{\textit{all teachers} $i=1,2,3,...,K$ in parallel}{
             $\Delta {\alpha}^{i} = \underset{\Delta{\alpha}^{i}}{\argmin}  \sum_{j=1}^{N_i} \ell^{*}(-\alpha^{t-1,i}_j-\Delta{\alpha}^{i}_j) + \frac{\lambda}{2}\|{\widetilde\theta}^{t-1} + \frac{1}{\lambda} \sum_{j=1}^{N_i}\Delta{\alpha}^{i}_{j} x^{i}_j y^{i}_j \|^2 + \lambda_{\alpha} \sum_{j=1}^{N_i} w^{i}_{j} |\alpha^{t-1,i}_j + \Delta{\alpha}^{i}_j|+\lambda_{\theta}\|{\widetilde\theta}^{t-1} + \frac{1}{\lambda} \sum_{j=1}^{N_i}\Delta{\alpha}^{i}_{j} x^{i}_j y^{i}_j - \theta^{*}\|^2$\\
             $\alpha^{t,[k]} = \Pi(\alpha^{t-1,[k]} + \frac{\beta_{K}}{K}\Delta{\alpha}^{[k]})$\\
      }
      
      \textit{Reduce on the central parameter server} ${\widetilde\theta}^{t} = {\widetilde\theta}^{t-1} + \frac{1}{\lambda}\sum_{i=1}^{K}\sum_{j=1}^{N_i} \alpha^{t,i}_{j} x^{i}_j$ \\
      \textit{Broadcast $\widetilde\theta^{t}$ to all $K$ teachers}
    }
\end{algorithm}
$\Pi$ is the projection operator to make the updated value of $\alpha^{(t),[k]}$ satisfy the box constraint. 
\subsubsection{Collaborative Teaching for Ridge Regression}
Given the feature $x^{i}_j$ and regression target $y^{i}_{j}$ of each data instance, we can define the objective of collaborative teaching for Ridge Regression similarly. 
\begin{equation}\label{rr_dual_regularized}
\small
\begin{split}
\alpha &= \underset{\alpha}{\argmin} \frac{1}{2}\sum_{i=1}^{K}\sum_{j=1}^{N_i} \ell^{*}(-\alpha^{i}_j) + \frac{1}{2\lambda}\|\sum_{i=1}^{K}\sum_{j=1}^{N_i}x^{i}_j \alpha^{i}_j\|^2 \\
& + \lambda_{\alpha}\sum_{i=1}^{K}\sum_{j=1}^{N_i} w^{i}_{j}|\alpha^{i}_j| + \lambda_{\theta} \|\theta^{*} - \frac{1}{\lambda} \sum_{i=1}^{K}\sum_{j=1}^{N_i} x^{i}_j \alpha^{i}_j\|^2 \\
\end{split}
\end{equation}
where $\ell^{*}(-\alpha^{i}_j) = \frac{1}{2} \|\alpha^{i}_j\|^2 - \alpha^{i}_{j} y^{i}_{j}$. It is thus easy to define collaboratively teaching ridge regression in algorithm~\ref{alg:fedML_rr}

\begin{algorithm}[t]
\caption{Block-Coordinate Descent for Collaborative Super Teaching of Ridge Regression}\label{alg:fedML_rr}
\textbf{Initialize}: $\alpha^{i}_{j} = 0$ for teachers and ${\widetilde\theta}^{0} = 0$\\
\For{$t=1$ \KwTo $T$}{ 
      \For{\textit{all teachers} $i=1,2,3,...,K$ in parallel}{
             $\Delta {\alpha}^{i} = \underset{\Delta{\alpha}^{i}}{\argmin} \sum_{j=1}^{N_i} \ell^{*}(-\alpha^{t-1,i}_j - \Delta\alpha^{i}_j) + \frac{\lambda}{2}\|\widetilde{\theta}^{t-1} + \frac{1}{\lambda}\sum_{j=1}^{N_i}x^{i}_j \Delta\alpha^{i}_j\|^2 + \lambda_{\alpha} \sum_{j=1}^{N_i} w^{i}_{j}|\alpha^{t-1,i}_j + \Delta{\alpha}^{i}_j|+  \lambda_{\theta}\|{\widetilde\theta}^{t-1} + \frac{1}{\lambda} \sum_{j=1}^{N_i}\Delta{\alpha}^{i}_{j} x^{i}_j - \theta^{*}\|^2$\\
             $\alpha^{t,i} = \alpha^{t-1,i} + \frac{\beta_{i}}{K}\Delta{\alpha}^{i}$\\
      }
      
      \textit{Reduce on the central parameter server} ${\widetilde\theta}^{t} = {\widetilde\theta}^{t-1} + \frac{1}{\lambda}\sum_{i=1}^{K}\sum_{j=1}^{N_i} \alpha^{t,i}_{j} x^{i}_j$ \\
      \textit{Broadcast $\widetilde\theta^{t}$ to all $K$ teachers}
    }
\end{algorithm}

\subsection{Computational complexity and communication cost}
As shown in algorithm~\ref{alg:fedML_lr} and algorithm~\ref{alg:fedML_rr},  estimating the incremental update of each block $\alpha^{i}$ is a convex quadratic programming problem. With appropriately set $\lambda_{\theta}$ and $\lambda_{\alpha}$, the quadratic programming problem is well scaled and can be solved in a well scalable way using polynomial time interior point algorithms, such as \cite{YinyuYe1989}. According to algorithm~\ref{alg:fedML}, only the step of aggregating the global variable $\widetilde\theta$ needs communication between the $K$ teachers and the learner. Assuming that in total $T$ iterations are needed in the block-coordinate descent in algorithm~\ref{alg:fedML}, the overall communication cost of running the collaborative teaching paradigm for both models is $O(TKd)$. In practices, $T=100$ is enough to achieve convergence of the block coordinate descent. 

\section{Experimental Study}\label{sec:exp}
\subsection{Experimental setup}
We test the proposed collaborative teaching method with both synthetic data set and real-world benchmark datasets (summarized in Table.\ref{tab:datasets}). For the synthetic classification and regression data set, we create clusters of random data instances following normal distribution. In the classification dataset, equal number of clusters are assigned to positive and negative classes to construct a balanced labelled data set. In the regression dataset, the regression target $Y$ is given by applying random linear regressor to $X$. The dimensionality of each data instance is fixed to $10$ universally. In the experimental study, we assume that each of the $K$ teachers hosts $\lfloor\frac{N}{K}\rfloor$ data instances as the local data set. To generate \textit{i.i.d.} data instances, the mean and variance of the normal distribution for data generation are kept the same for different teachers. The summary of the real-world datasets is shown by Table.\ref{tab:datasets}, which are used to evaluate practical performances of the proposed method over large-scale real-world data samples. The empirical study over the real-world data samples uniformly the data set and assign $\lfloor\frac{N}{K}\rfloor$ instances to each teacher. 

To generate the target of teaching in the study, we run standard LR and RR on all the data samples hosted by $K$ teachers to derive true model parameter $\theta^{gt}$. The teaching target $\theta^{*}$ is given by adding a white Gaussian noise $\tau \in R^{d}$, as $\theta^{*} = \theta^{gt} + \tau$. We fix the magnitude of $\tau$ as the same of that of $\theta^{gt}$ in the following experiments to generate a teaching target with reasonable difficulty. To measure the teaching performances, we use the teaching risk $R^{*}$ as the major metric. In addition, in the binary classification scenario, we apply both the teaching target $\theta^{*}$ and the learned parameter $\hat\theta_S$ on the whole data set. We count the fraction of the data instances where the output labels of the teaching target and the learned model are consistent. The higher the fraction value is, the better the teaching performance is, as the goal of teaching is to approximate the target model as close as possible. Similarly for regression, we measure $r$-square score between the regression output of the teaching target model and the learned model on the whole synthetic regression data set as the metric measuring the teaching quality for regression. The two additional metrics are noted as ${\rho}_{lr}$ and $\rho_{rr}$ in the experiments.

\begin{table}[]
\centering
\small
\caption{Summary of public real-world benchmark datasets.}
\label{tab:datasets}
\resizebox{\linewidth}{!} {
\begin{tabular}{|c|c|c|}
\hline
\textbf{Dataset} & \textbf{No. of Instances} & \textbf{No. of Features} \\ \hline
     Higgs   &        1,000,000          &       28            \\ \hline
     Superconduct & 21,263 & 81 \\\hline
\end{tabular}
}
\end{table}

We compare the proposed collaborative teaching method to the primitive oblivious teaching strategy. To organize the oblivious teaching, we further require that each teacher selects $\frac{|S|}{K}$ instances as the identified local subset, as there is no heuristic preference over any specific teacher. The oblivious teaching is conducted by running the MINLP based teaching paradigm \cite{ma2018teacher} on each teacher. The selected data instances are aggregated to form the training set of the learner. The proposed collaborative teaching method is implemented with Spark \textit{TFOCS} library on a AWS EC2 public cloud server, with one core per teacher. For implementing the oblivious teaching method, it is difficult to find an open-sourced \textit{MINLP} library tailored for parallel computing environments. We thus use Spark to call the MINLP solver of \textit{NEOS} \cite{dolan_2001} for each teacher and aggregate the selected data instances to form the learner's training set. We record the running time to evaluate and compare the scalability of both teaching methods, as indicated by $\kappa$ in the empirical study.

\subsection{Benchmark with synthetic classification and regression datasets}
For the tests of both classification and regression scenarios, we vary the total number $N$ of synthetic data instances as 5000, 10000, 50000, 100000 and 500000 to cover intermediate and large-scale data volumes. For each choice of $N$, we further set the number of teachers $K$ to be 5 and 10 respectively. For a fixed combination of $N$ and $K$, we run 10 trials. In each trial, we draw randomly an \textit{\textit{iid}} synthetic instances and apply the proposed method of collaborative super teaching. We show the fraction of the selected subset $|S|/N$ that achieves the minimum average teaching risk of the 10 trials in Table.\ref{tab:lr} and Table.\ref{tab:rr}. $\lambda_{\alpha}$ and $\lambda_{\theta}$ are the parameters of the proposed collaborative teaching method. In the experimental study of both the classification and regression scenario, both parameters are tuned empirically using validation data instances that are generated independently besides from the benchmark set. It is interesting to find out that the values of $\lambda_{\alpha}$ and $\lambda_{\theta}$ are insensitive to the varying $N$ and $K$. Therefore, we fix $\lambda_{\alpha}$ as 0.1 and $\lambda_{\theta}$ as 1000 in the binary classification scenario. In the regression scenario, they are fixed as 1 and 2000 respectively. We also run the MINLP based teaching paradigm as a centralized teacher over all the training data instances,  as indicated by \textit{MINLP} in both tables. We compare to the centralized teaching to highlight the computing efficiency of the proposed teaching paradigm. 


\begin{table*}
\small
\centering
\caption{Comparison of the teaching performance in the binary classification scenario}
\label{tab:lr}
\resizebox{\linewidth}{!} {
\begin{tabular}{|c|c|c|c|c|c|c|c|c|c|c|c|c|c|}
\hline
\multirow{2}{*}{\textbf{N}} & \multirow{2}{*}{\textbf{K}}  & \multicolumn{4}{|c|}{\textbf{Collaborative Super Teaching}}  & \multicolumn{4}{|c|}{Oblivious Super Teaching} & \multicolumn{4}{|c|}{MINLP}  \\\cline{3-14}
 &  & $|S|/N$ & $R^{*}(\hat\theta_S)$ & $\rho_{lr}$ & $\kappa$ & $|S|/N$ & $R^{*}(\hat\theta_S)$ & $\rho_{lr}$ & $\kappa$ & $|S|/N$ & $R^{*}(\hat\theta_S)$ & $\rho_{lr}$ & $\kappa$\\\hline
\multirow{2}{*}{5000} &5 & \textbf{3.7e-2}  & \textbf{0.43} & \textbf{0.95} & \textbf{7.15s} &  6.0e-2 & 0.54 & 0.90 & 200.53s &\multirow{2}{*}{4.0e-2} & \multirow{2}{*}{0.30} & \multirow{2}{*}{0.97} & \multirow{2}{*}{175.91s}  \\\cline{2-10}
 & 10 & \textbf{4.5e-2} & \textbf{0.37} & \textbf{0.92} & \textbf{8.13s} & 5.8e-2 & 0.62 & 0.87 & 197.33s &  & &  &    \\\hline
\multirow{2}{*}{10000} &5 & \textbf{1.5e-2}  & \textbf{0.23} & \textbf{0.94} & \textbf{16.90s} & 2.0e-2 & 0.63 & 0.87 & 320.69s &\multirow{2}{*}{N/A} & \multirow{2}{*}{N/A} & \multirow{2}{*}{N/A} & \multirow{2}{*}{N/A} \\\cline{2-10}
 & 10 & \textbf{1.0e-2} & \textbf{0.26} & \textbf{0.97} & \textbf{18.85s} & 2.0e-2 & 0.54 & 0.89 & 327.16s &  & &  &   \\\hline
\multirow{2}{*}{50000} &5 & \textbf{2.5e-2}  & \textbf{0.35} & \textbf{0.93} & \textbf{67.20s} & N/A & N/A  & N/A & N/A & \multirow{2}{*}{N/A} & \multirow{2}{*}{N/A} & \multirow{2}{*}{N/A} & \multirow{2}{*}{N/A}   \\\cline{2-10}
 & 10 & \textbf{2.4e-2} & \textbf{0.36} & \textbf{0.96} & \textbf{72.52s} & 1.3e-2  & 0.22 & 0.94  & 2100s &  & &  &   \\\hline
\multirow{2}{*}{100000} & 5 & \textbf{2.9e-2}  & \textbf{0.33} & \textbf{0.95} & \textbf{180.24s} & N/A  & N/A  & N/A  & N/A & \multirow{2}{*}{N/A} & \multirow{2}{*}{N/A} & \multirow{2}{*}{N/A} & \multirow{2}{*}{N/A}\\\cline{2-10}
 & 10 & \textbf{3.0e-2} & \textbf{0.41} & \textbf{0.93} & \textbf{179.12s} & N/A  & N/A  & N/A & N/A & & &  &\\\hline
\multirow{2}{*}{500000} & 5 & \textbf{6.4e-2}  & \textbf{0.15} & \textbf{0.98} & \textbf{1064.15s} & N/A  & N/A  & N/A  & N/A & \multirow{2}{*}{N/A} & \multirow{2}{*}{N/A} & \multirow{2}{*}{N/A} & \multirow{2}{*}{N/A}  \\\cline{2-10}
 & 10 & \textbf{7.2e-2} & \textbf{0.12} & \textbf{0.98} & \textbf{1100.75s}& N/A  & N/A  & N/A & N/A  & & &  &\\\hline

\end{tabular}
}
\end{table*}

\begin{table*}
\small
\centering
\caption{Comparison of the teaching performance in the regression scenario}
\label{tab:rr}
\resizebox{\linewidth}{!} {
\begin{tabular}{|c|c|c|c|c|c|c|c|c|c|c|c|c|c|}
\hline
\multirow{2}{*}{\textbf{N}} & \multirow{2}{*}{\textbf{k}}  & \multicolumn{4}{|c|}{\textbf{Collaborative Super Teaching}}  & \multicolumn{4}{|c|}{Oblivious Super Teaching} & \multicolumn{4}{|c|}{MINLP}  \\\cline{3-14}
 &  & $|S|/N$ & $R^{*}(\hat\theta_S)$ & $\rho_{lr}$ & $\kappa$ & $|S|/N$ & $R^{*}(\hat\theta_S)$ & $\rho_{lr}$ & $\kappa$ & $|S|/N$ & $R^{*}(\hat\theta_S)$ & $\rho_{lr}$ & $\kappa$\\\hline
\multirow{2}{*}{5000} &5 & \textbf{1.20e-1}  & \textbf{81.53} & \textbf{0.82} & \textbf{2.04s} & 2.00e-1 & 112.93 & 0.71 & 230.53s &\multirow{2}{*}{1.20e-1} & \multirow{2}{*}{98.09} & \multirow{2}{*}{0.76} & \multirow{2}{*}{195.84s}  \\\cline{2-10}
 & 10 & \textbf{1.12e-1} & \textbf{80.43} & \textbf{0.83} & \textbf{2.38s} & 1.80e-1 & 120.54 & 0.70 & 262.65s &  & &  &    \\\hline
\multirow{2}{*}{10000} &5 & \textbf{9.00e-2}  & \textbf{69.52} & \textbf{0.86} & \textbf{4.25s} & 2.00e-1 & 93.20 & 0.75 & 320.28s &\multirow{2}{*}{2.00e-1} & \multirow{2}{*}{87.96} & \multirow{2}{*}{0.76} & \multirow{2}{*}{506.01s} \\\cline{2-10}
 & 10 & \textbf{7.00e-2} & \textbf{67.46} & \textbf{0.87} & \textbf{5.15s} & 2.50e-1 & 94.20 & 0.76 & 570.32s &  & &  &   \\\hline
\multirow{2}{*}{50000} &5 & \textbf{1.00e-1}  & \textbf{110.62} & \textbf{0.84} & \textbf{35.16s} & N/A & N/A  & N/A & N/A & \multirow{2}{*}{N/A} & \multirow{2}{*}{N/A} & \multirow{2}{*}{N/A} & \multirow{2}{*}{N/A}   \\\cline{2-10}
 & 10 & \textbf{1.12e-1} & \textbf{118.36} & \textbf{0.82} & \textbf{32.53s} & N/A  & N/A & N/A  & N/A &  & &  &   \\\hline
\multirow{2}{*}{100000} & 5 & \textbf{3.00e-1}  & \textbf{99.85} & \textbf{0.88} & \textbf{81.80s} & N/A  & N/A  & N/A  & N/A & \multirow{2}{*}{N/A} & \multirow{2}{*}{N/A} & \multirow{2}{*}{N/A} & \multirow{2}{*}{N/A}\\\cline{2-10}
 & 10 & \textbf{3.00e-1} & \textbf{101.36} & \textbf{0.88} & \textbf{101.03s} & N/A  & N/A  & N/A & N/A & & &  &\\\hline
\multirow{2}{*}{500000} & 5 & \textbf{3.60e-1}  & \textbf{51.72} & \textbf{0.94} & \textbf{310.42s} & N/A  & N/A  & N/A  & N/A & \multirow{2}{*}{N/A} & \multirow{2}{*}{N/A} & \multirow{2}{*}{N/A} & \multirow{2}{*}{N/A}  \\\cline{2-10}
 & 10 & \textbf{3.60e-1} & \textbf{49.67} & \textbf{0.95} & \textbf{395.91s}& N/A  & N/A  & N/A & N/A  & & &  &\\\hline

\end{tabular}
}
\end{table*}

The collaborative super teaching method selects less than 0.1 and 0.4 of the data instances to achieve accurate teaching result in both the classification and the regression test. Given $N$ fixed, increasing $K$ barely changes the teaching risk and the decision consistency between the target model and the model learned with the selected subset. However, it slightly increases the running time due to the increased communication cost during the global aggregation and broadcasting of ${\widetilde\theta}^{t}$. In all of the tests, the collaborative super teaching method runs for 85 to 150 iterations to reach convergences. With the same $N$, more teachers (larger $K$) requires more iterations before convergence. On one hand, collaborating with more teachers leads to smaller block size of the block coordinate descent, which causes slower convergence \cite{ABeck2013}. On the other hand, more teachers help to reduce the computational cost on each teacher. Depending on the computational resource budget of the teachers, we can benefit from the balance to organize efficient collaboration of the teachers. In general, compared to the oblivious teaching and the centralized teaching method, the collaborative teaching method provides significantly lower or similar teaching risk and better approximate the target model with the selected training subset in both tests. It requires distinctively less running time. The collaborative teaching method costs less than $5\%$ of the running time compared to both of the opponents according to Table.\ref{tab:lr} and Table.\ref{tab:rr}. Notably, the time and storage cost of the oblivious teaching becomes prohibitively expensive when $N \geq {10000}$. NEOS can't get results given the large $N$. We write $N/A$ in case NEOS fails to solve the MINLP problem. In contrast, the proposed collaborative teaching method can still produce accurate teaching output with acceptable time cost. Despite of the difference of  implementation details of the three teaching methods, the difference of running time confirms the superior computational efficiency of the proposed collaborative teaching paradigm. Benefited from the consensus optimization process, the proposed collaborative teaching paradigm provides a highly scalable solver to the distributed super teaching task. Moreover, it is interesting to find out the centralized teaching paradigm performs better than the oblivious teaching method. This observation is consistent with what Theorem.\ref{thm:kparty} unveils: information sharing is the key to achieve good teaching cooperation. Stand-alone teaching without inter-communication between teachers can do harm to the teaching performance. 
 
\subsection{Benchmark with real-world data sets}
\begin{figure}[t]
    \subfloat[Higgs Data ] {
        \raisebox{-\height}{\includegraphics[width=0.45\linewidth]{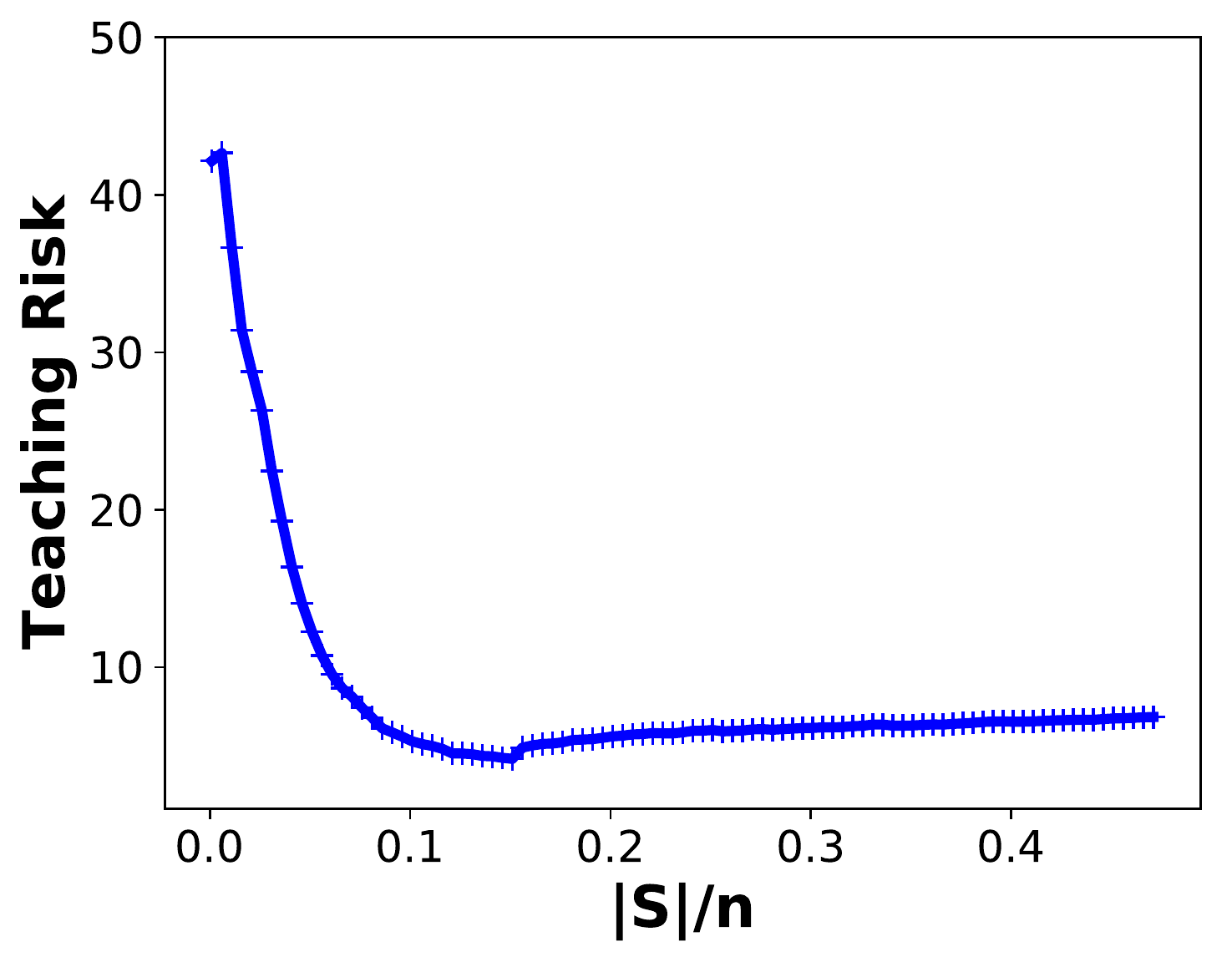}\label{fig:higgs_teachingrisk}}
    }
    \hfill
    \subfloat[Superconduct Data]{
        \raisebox{-\height}{\includegraphics[width=0.45\linewidth]{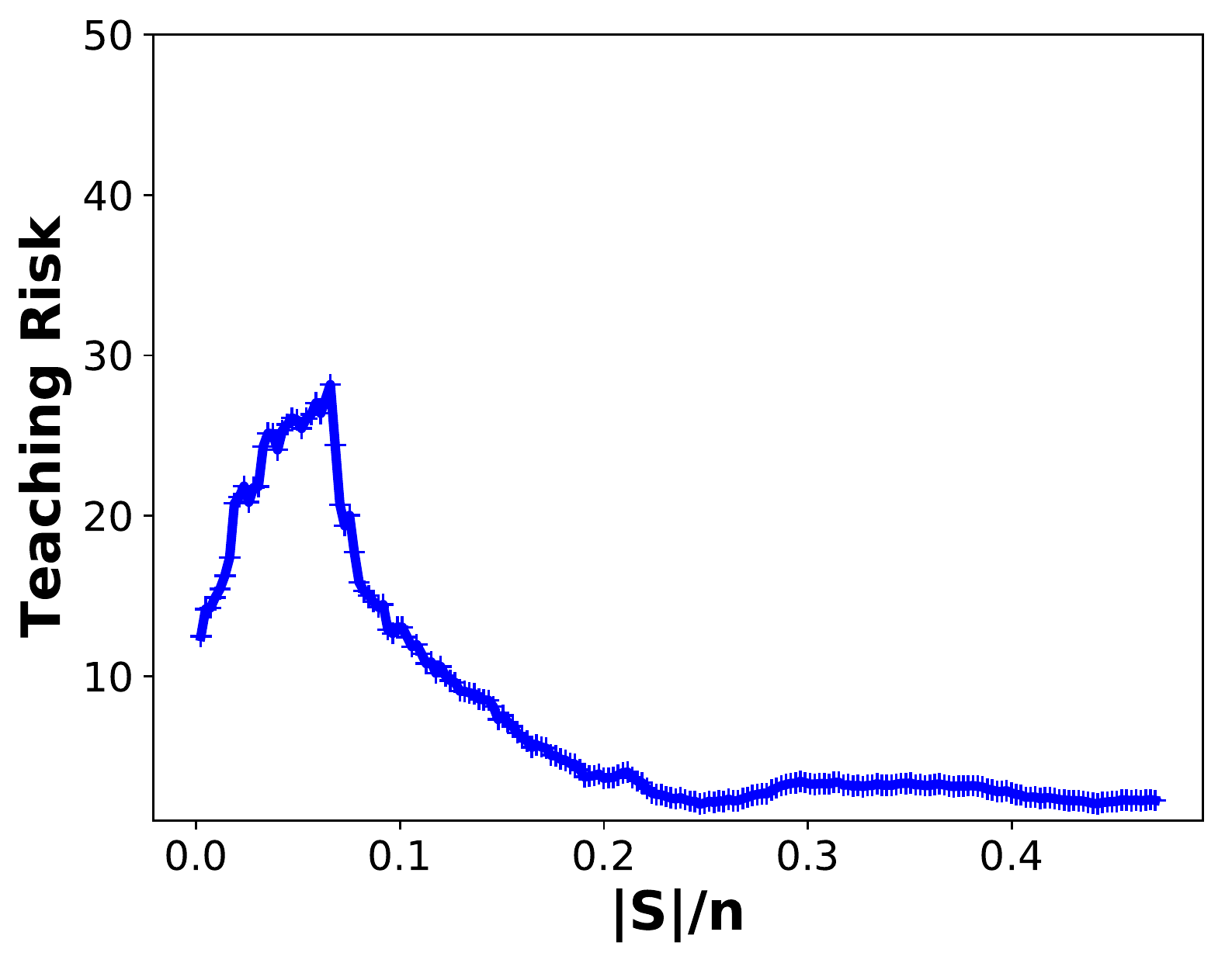}\label{fig:superconduct_teachingrisk}}
    }
    \hfill
    \caption{Teaching risk variation with different number of the selected data instances}
	\label{fig:teachingrisk}
   \vspace{-0.5cm}
\end{figure}

\begin{figure}[t]
    \subfloat[Higgs Data ] {
        \raisebox{-\height}{\includegraphics[width=0.45\linewidth]{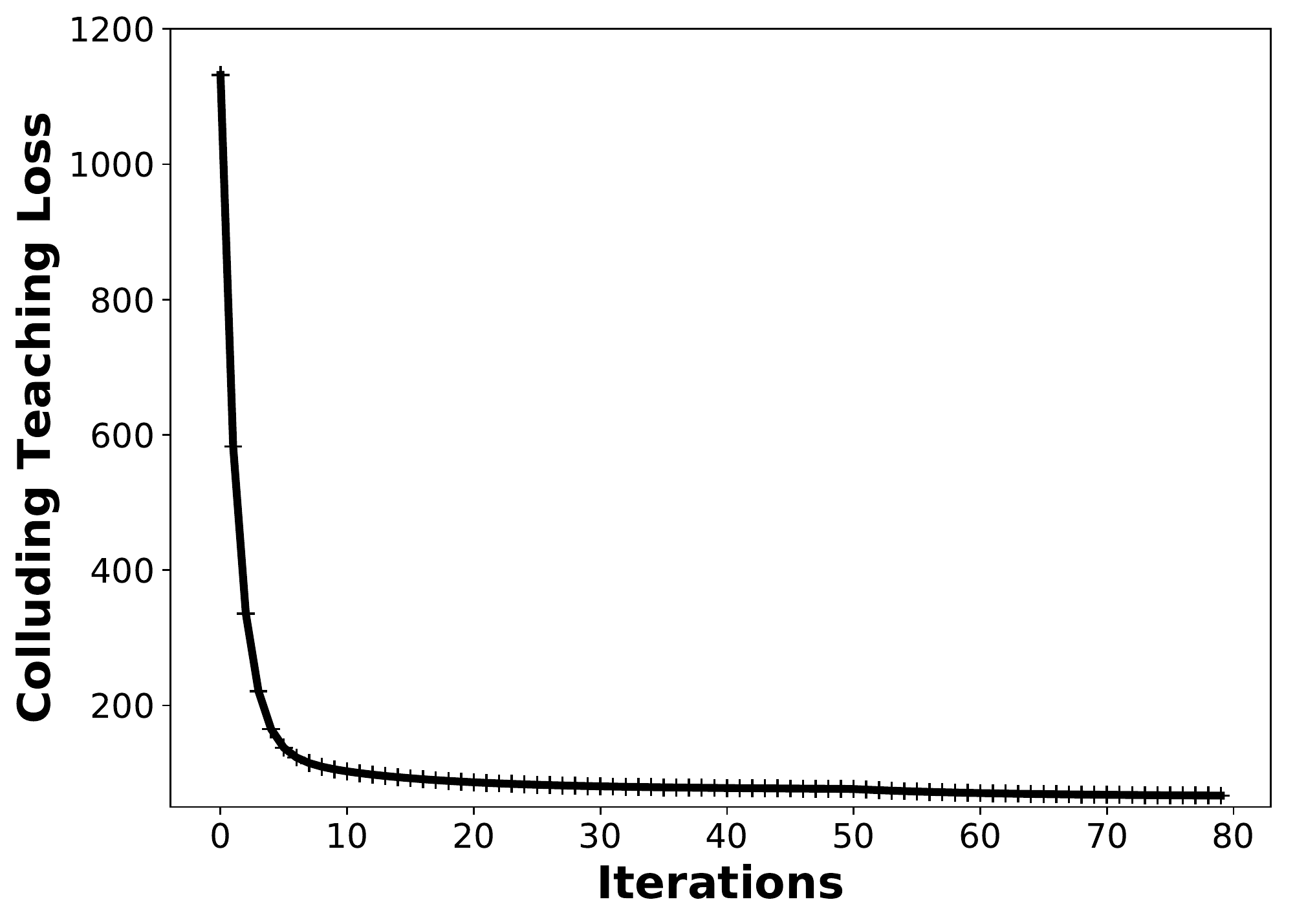}\label{fig:higgs_obj}}
    }
    \hfill
    \subfloat[Superconduct Data]{
        \raisebox{-\height}{\includegraphics[width=0.45\linewidth]{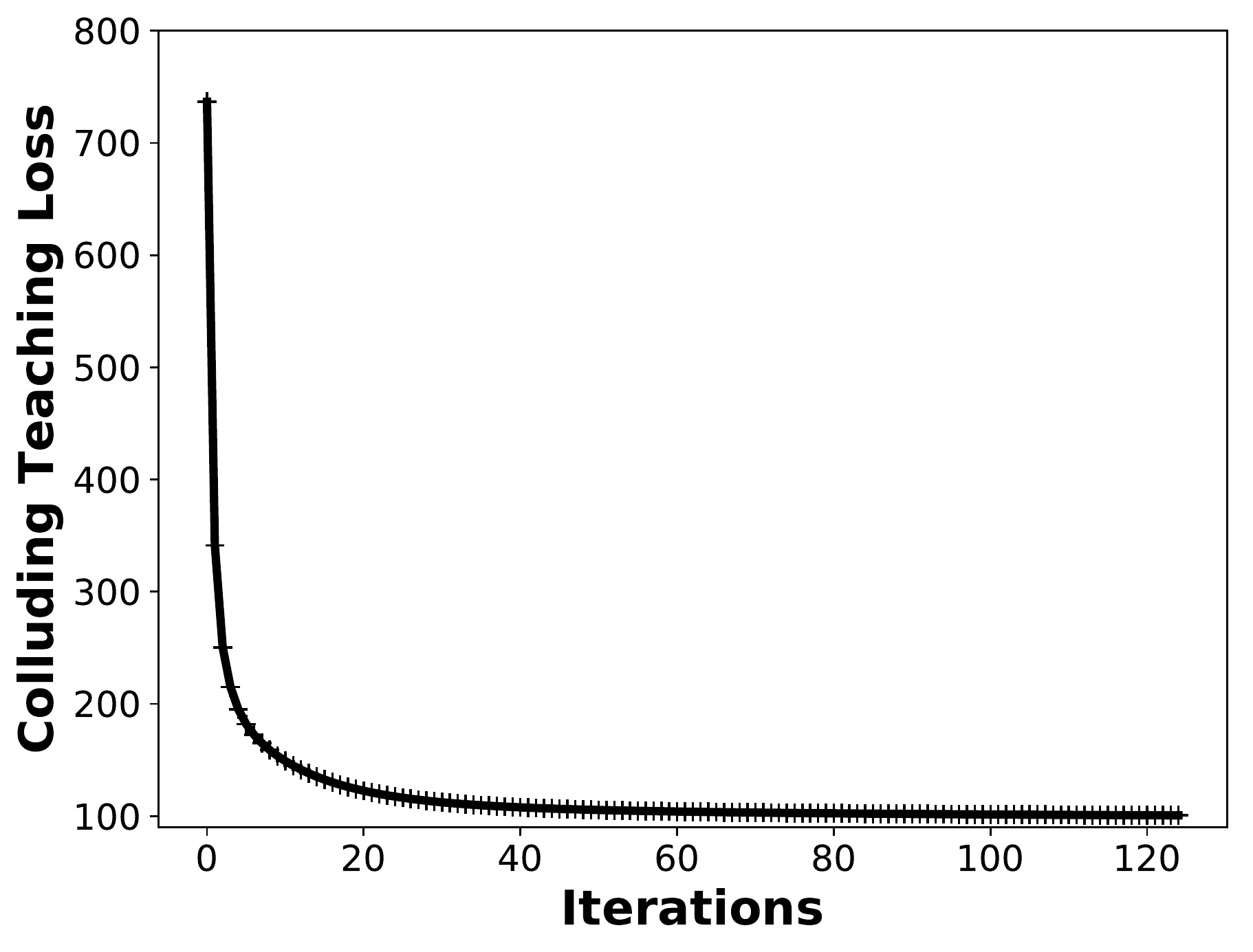}\label{fig:superconduct_obj}}
    }
    \hfill
    \caption{Convergence of the quadratic programming based collaborative teaching process}
	\label{fig:objectivefunc}
   \vspace{-0.5cm}
\end{figure}

Two real-world data sets, \textit{Higgs} and \textit{Superconduct}, are employed to test the collaborative super teaching method for $l_2$-regularized Logistic Regression and Ridge Regression respectively. The number of the teachers is chosen to be 5 on both data sets. The setting of $\lambda_{\alpha}$, $\lambda_{\theta}$, the computing platform and the teaching goal follow the same setting as the test on synthetic data. For Higgs data set, we randomly sample 1000000 instances from the whole set for 10 times and re-run the proposed method on the sampled Higgs data samples. Figure~\ref{fig:higgs_teachingrisk} illustrates the variation of the averaged teaching risk of the proposed collaborative teaching method by incrementally increasing the number of the jointly selected instances. In Figure~\ref{fig:higgs_obj}, we demonstrate how the objective function value of the proposed teaching method diminishes as the iterative block-coordinate descent runs. On Higgs data set, the proposed method selects only 15\% of the 1000000 instances to achieve the teaching risk of 3.28. The corresponding consistency score $\rho_{lr}$ is 0.99. It indicates that the learner manages to approximate the expected target model perfectly with the selected subsets given the teachers. Interestingly, the teaching risk declines at first as ${|S|}/{N}$ increases to $15\%$. After this turning point, the teaching risk begins to increase again. The observation is consistent with our intuitive understanding about the teaching process. Insufficient and too many data instances can do harm equally to the teaching performances.  From Figure~\ref{fig:higgs_obj}, we can find the objective function value of the proposed collaborative teaching method declines rapidly within 50 iterations. In this experiment, the consensus optimization process of the proposed collaborative teaching paradigm converges with 80 iteration steps. which costs 2095.48s. Similarly, Figure~\ref{fig:superconduct_teachingrisk} shows the declination of teaching risk by increasing gradually the number of selected instances on \textit{Superconduct} data set. As seen in the figure, the proposed teaching method selects 25\% of the data instances to reach the teaching risk of 2.23 and $\rho_{rr}$ of 0.97. Figure~\ref{fig:superconduct_obj} illustrates the declination of the objective functions values on \textit{Superconduct} achieves. Similar pattern of the teaching risk variation is witnessed in Figure~\ref{fig:superconduct_teachingrisk}, compared to Figure~\ref{fig:higgs_teachingrisk}. The turning point of the teaching risk curve confirms empirically the existence of the optimal subset for teaching. Based on the selected subset, the learner can accurately fit the target regression model. Minimizing the collaborative teaching objective for Ridge Regression converges within 125 iterations, which costs only 35.12s. 

\section{Conclusion and Discussion}\label{sec:conclusion}
In this paper, we explore how to organize scalable, collaborative, and privacy-preserving super teaching with multiple teachers. We formulate a distributed convex optimization problem for conducting consensus super teaching with varying number of teachers, and adopt a block descent based solver to optimize each teacher's selection on teaching items. Our approach preserves data privacy during the collaborative teaching process. We show that the proposed collaborative teaching scheme can achieve lower teaching risk than the non-collaborative scheme. 
Empirical results on both synthetic and real-word data sets confirm the superior performance of the proposed collaborative teaching method over the non-collaborative solution. Future work will study practical use of distributed and privacy-preserving super teaching based on the proposed collaborative teaching framework, e.g., we plan to explore the teaching goals that are realistic to practical use, such as AUC-maximization oriented goals.

\bibliography{ijcnn}
\bibliographystyle{IEEEtran}

\end{document}